\documentclass[10pt, journal]{IEEEtran}

\usepackage{amsmath,amssymb,amsfonts}
\usepackage{graphicx}
\usepackage{pifont}
\usepackage[edges]{forest}
\usepackage{tikz}
\usetikzlibrary{trees}

\usepackage{amsthm}
\usepackage[figuresright]{rotating}

\usepackage{graphicx}
\graphicspath{{/}{fig/}}

\usepackage{array}
\usepackage{textcomp}
\usepackage{multirow}
\usepackage[table]{xcolor}
\usepackage{booktabs}
\usepackage{pifont}

\usepackage{mathtools}
\usepackage{breqn}
\usepackage{float}

\usepackage{pgfplots}
\pgfplotsset{compat=1.7}
\usepgfplotslibrary{groupplots}

\usepackage{cite} 

\usepackage{caption}
\usepackage{subcaption}

\usepackage{multirow,tabularx}
\usepackage{hyperref}
\usepackage{flushend}
\usepackage{algorithmic}
\usepackage[vlined, ruled, shortend]{algorithm2e}
\usepackage{enumitem}
\usepackage{cuted}    
\usepackage{makecell}

\usepackage{soul} 

\usepackage{wrapfig} 
\usepackage{adjustbox} 
\usepackage{makecell} 

\usepackage[usenames,dvipsnames]{xcolor}

\usepackage{tikz}
\tikzset{%
    baseline,
    inner sep=2pt,
    minimum height=12pt,
    rounded corners=2pt  
}
\newcommand{\code}[1]{\mbox{
    \ttfamily
    \tikz \node[anchor=base,fill=black!12]{#1};
}}

\usepackage{cuted}

\usepackage{overpic}
\usepackage{listings}
\usepackage{xcolor}
\usepackage{tcolorbox}
\definecolor{jsonbg}{HTML}{F7F9FC}   
\definecolor{jsonframe}{HTML}{A0A0A0} 
\definecolor{jsonstring}{HTML}{228B22} 
\definecolor{jsonkey}{HTML}{0000FF}    
\definecolor{jsonnumber}{HTML}{B22222} 

\tcbuselibrary{listings, breakable}

\lstdefinestyle{json}{
    basicstyle=\ttfamily\footnotesize,
    numbers=none,
    backgroundcolor=\color{jsonbg},
    showstringspaces=false,
    breaklines=true,
    frame=single,
    rulecolor=\color{jsonframe},
    stringstyle=\color{jsonstring},
    keywordstyle=\color{jsonkey},
    numberstyle=\color{jsonnumber},
    literate=
     *{true}{{{\color{jsonnumber}true}}}{4}
      {null}{{{\color{jsonnumber}null}}}{4},
}

\tcbset{
  mybox/.style={
    colback=gray!10,        
    colframe=blue!80,      
    fonttitle=\bfseries,
    boxrule=0.5pt,
    arc=2pt,
    outer arc=2pt,
    left=6pt,
    right=6pt,
    top=6pt,
    bottom=6pt,
    breakable
  }
}

\newlength\figureheight
\newlength\figurewidth
\SetAlCapNameFnt{\footnotesize}
\SetAlCapFnt{\footnotesize}

\title{
    ROSBag MCP Server: \\
    Analyzing Robot Data with LLMs for \\
    Agentic Embodied AI Applications
}

\author{
    Lei Fu\textsuperscript{*1},
    Sahar Salimpour\textsuperscript{*2}, Leonardo Militano\textsuperscript{1}, 
    Harry Edelman\textsuperscript{4}, \\
    Jorge Peña Queralta\textsuperscript{3,4}, Giovanni Toffetti\textsuperscript{1}\\[+1em]
    
    \textsuperscript{1}\href{https://www.zhaw.ch/en/engineering/institutes-centres/init}{Institute of Computer Science, Zurich University of Applied Sciences} \\[+.42em]
    \textsuperscript{2}\href{https://www.utu.fi/en/university/faculty-of-technology/computing}{Department of Computing, University of Turku, Finland} \\[+.42em]
    \textsuperscript{3}\href{https://www.zhaw.ch/en/engineering/institutes-centres/cai}{Centre for Artificial Intelligence, Zurich University of Applied Sciences} \\[+.42em]
    \textsuperscript{4}\href{https://binabik.ai}{Binabik.ai}

    \thanks{\textsuperscript{*}Authors with equal contribution.}

    \thanks{Emails:\{fule, milt, penq, toff\}@zhaw.ch,  sahars@utu.fi, harry@binabik.ai}
}

\begin{document}

\maketitle

\thispagestyle{plain}
\pagestyle{plain}

\begin{abstract}
    \label{sec:abstract}
    Agentic AI systems and Physical or Embodied AI systems have been two key research verticals at the forefront of Artificial Intelligence and Robotics, with Model Context Protocol (MCP) increasingly becoming a key component and enabler of agentic applications. However, the literature at the intersection of these verticals, i.e., Agentic Embodied AI, remains scarce. This paper introduces an MCP server for analyzing ROS and ROS\,2 bags, allowing for analyzing, visualizing and processing robot data with natural language through LLMs and VLMs. We describe specific tooling built with robotics domain knowledge, with our initial release focused on mobile robotics and supporting natively the analysis of trajectories, laser scan data, transforms, or time series data. This is in addition to providing an interface to standard ROS\,2 CLI tools (\textit{ros2 bag list} or \textit{ros2 bag info}), as well as the ability to filter bags with a subset of topics or trimmed in time. Coupled with the MCP server, we provide a lightweight UI that allows the benchmarking of the tooling with different LLMs, both proprietary (Anthropic, OpenAI) and open-source (through Groq). Our experimental results include the analysis of tool calling capabilities of eight different state-of-the-art LLM/VLM models, both proprietary and open-source, large and small. Our experiments indicate that there is a large divide in tool calling capabilities, with Kimi K2 and Claude Sonnet 4 demonstrating clearly superior performance. We also conclude that there are multiple factors affecting the success rates, from the tool description schema to the number of arguments, as well as the number of tools available to the models. The code is available with a permissive license at \url{https://github.com/binabik-ai/mcp-rosbags}.
\end{abstract}

\IEEEpeerreviewmaketitle


\section{Introduction}

The development of autonomous robots capable of interacting with humans through natural language has been a long-standing objective in robotics research~\cite{zeng2023large}. These capabilities include commanding and controlling the robot, as well as interpreting and analyzing its state and diverse streams of sensor data via conversational interfaces, all without requiring expert-level knowledge in robotics or programming~\cite{wang2024multimodal, kawaharazuka2025vision}. Notably, large language models (LLMs) and, building on them, Vision-Language Models (VLMs) have demonstrated considerable promise in bridging the gap between human intent and robotic execution by translating natural language instructions into structured actions~\cite{wang2025large,zitkovich2023rt}. Consequently, there has been a growing interest in embodied AI agent systems, capable of reasoning, planning, maintaining memory, and interacting with humans~\cite{salimpour2025towards}. A significant portion of robotic systems relies on the Robot Operating System (ROS), and recent works, such as ROSA~\cite{royce2025enabling} and RAI~\cite{rachwal2025rai}, have explored LLM-based agents that bridge the gap between natural language and robot stack.

These agents often rely on LLMs augmented with the ability to integrate with external APIs. The Model Context Protocol (MCP), introduced by Anthropic in late 2024~\cite{anthropic2024mcp}, is an open interoperability standard designed to simplify and unify the way LLM models connect with external tools and replace the custom-built API integrations that currently hinder scalability and integration.

ROSbag files are a standardized format for storing large volumes of time-synchronized robotic data from diverse sensors. Analyzing these datasets is essential for debugging, performance evaluation, and extracting insights from robotic systems. However, direct analysis of this data using state-of-the-art conversational AI models is challenging, as they cannot natively parse the binary format, often exceed context length limits, and lack direct access to structured metadata and sensor semantics, leading to incomplete or inaccurate results. To address these limitations, this paper presents an MCP server that enables comprehensive and efficient human interaction with ROSbag files, facilitating rapid and in-depth analysis of recorded datasets.

\begin{figure*}
    \centering
    \includegraphics[width=.98\textwidth]{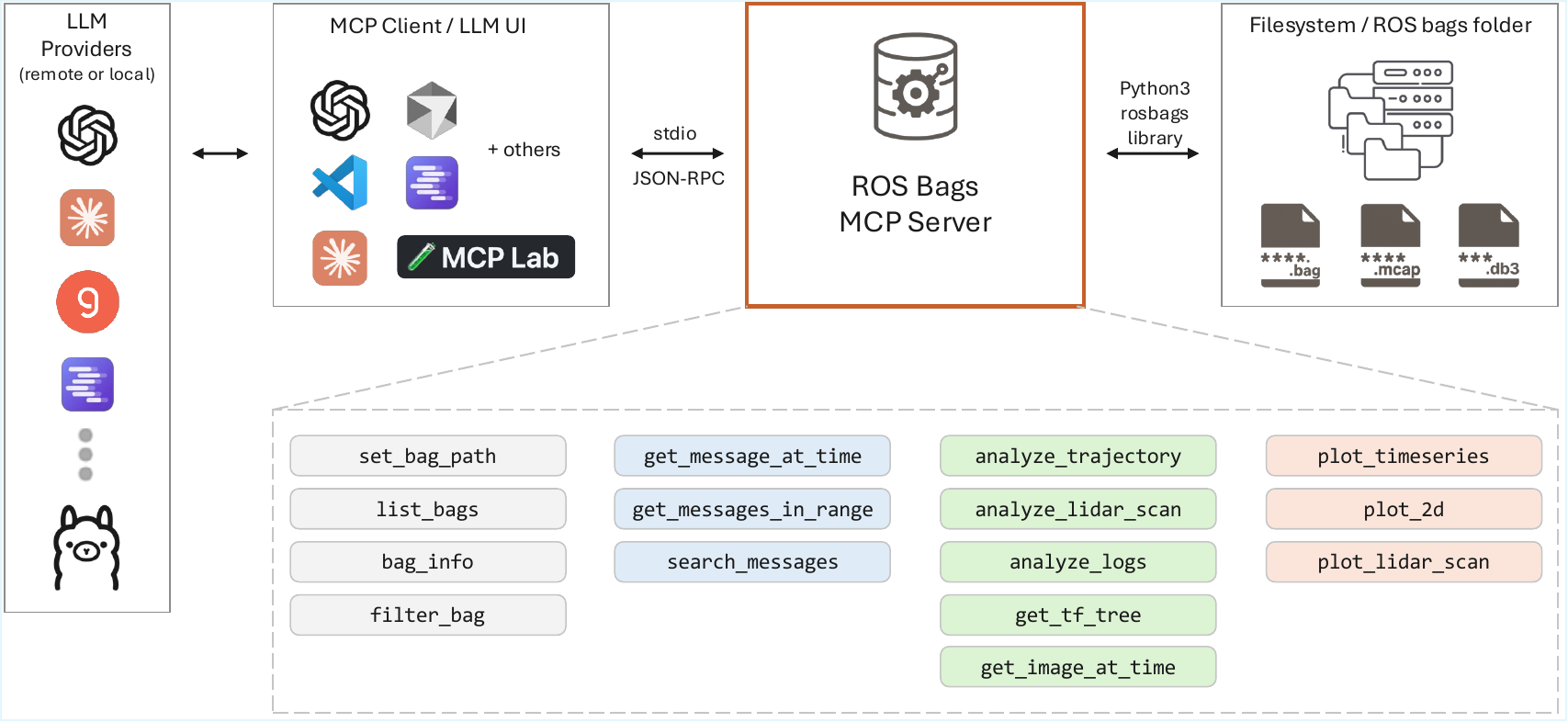}
    \caption{{The overview of the agentic chain of the ROSbag MCP Server. The diagram illustrates the different components, including the available tools in the MCP server, the capability to read various ROSbag formats from the filesystem, and the overall data flow between the MCP client and server. The MCP Host represents the runtime environment that mediates communication between the LLM client (e.g., Claude Desktop or MCP Lab) and the MCP server, managing JSON-RPC connections to the selected LLM provider.}}
    \label{fig:diagram}
\end{figure*}

The main contribution of this paper is the design and development of \textit{rosbags-mcp} framework. Unlike previous MCP integrations, our work focuses specifically on enabling LLM/VLM-driven data analysis of ROSbag files. To the best of our knowledge, this is the first paper to propose an MCP interface for analyzing and processing ROS native data. Compared with existing open-source tooling, we introduce a wider and more comprehensive toolset, grounded on robotics domain knowledge. Additionally, we are the first to benchmark tool calling capabilities of different proprietary and open-source LLMs and VLMs within a robotics data context. This is particularly relevant due to the increasing complexity of the proposed tools, in comparison to MCP servers in other domains that act mainly as translators. To enable such benchmarking, we also develop and introduce a web UI, \textit{MCP Lab}, which provides built-in visualization of plots generated by our MCP server. Figure~\ref{fig:diagram} illustrates the components and data flow of the proposed tool within the agentic system. 

The remainder of the paper is organized as follows. Section II reviews related work on agentic AI applications in robotics, recent advances in ROSbag analytics, and the development of MCP servers within the ROS community. Section III presents the implementation of our proposed ROSbag analytics MCP server, focusing on the technical aspects of MCP and robotics data analysis tools. Section IV reports the experimental results, and Section V concludes the paper.

\section{Related Works}

The literature in integrating the MCP protocol with ROS and ROS\,2 systems for natural language interfaces is scarce, due to the young age of MCP, proposed only in late 2024. The field of Agentic AI in general is also progressing rapidly, with Agentic Embodied AI in particular being a nascent research area. Therefore, in this section we focus on covering the literature from the general viewpoint of integrating LLMs and VLMs in robotics. We also cover existing tooling for analyzing ROSbags, and community-led MCP servers for ROS.

\subsection{ROS-based AI Agents}
Robot systems are often modular, allowing easy integration and modification of processes. The widely used ROS offers ready-to-use packages and libraries for many common robot systems. Limited yet promising developments have emerged in creating ROS-based robotic AI agents that bridge the gap between LLM-based reasoning and robotic middleware~\cite{salimpour2025towards}. A notable implementation is the ROS-LLM framework, which exposes ROS actions and services as atomic action tools to the LLM, bridging the gap between natural language commands and robot control for non-experts~\cite{mower2024ros}. ROSA~\cite{royce2025enabling} is an LLM-based agent built on the LangChain framework and the ReAct agent paradigm. Its core innovation lies in abstracting ROS operations into tool-enabled Python functions. The NASA JPL implementation of ROSA serves as an AI-powered assistant for both ROS1 and ROS2 systems, helping developers interact with robots using natural language queries, making robotics development more accessible and efficient. RAI~\cite{rachwal2025rai} is another approach in this field, serving as a flexible embodied multi-agent framework designed to integrate LLM reasoning with robotic systems (e.g., ROS 2).

Another example is \textit{ros2ai}~\cite{ros2ai}, which translates a generic user prompt into a specific ROS\,2 command-line interface call. While some interaction is possible, it remains limited, and the process is largely unidirectional. An early work in this area was also Microsoft's ChatGPT for robotics integration~\cite{vemprala2024chatgpt}. Such LLM-powered robot learning (as agentic applications) solutions are emerging as opposed to other research areas in RL~\cite{salimpour2025sim, zhao2020sim} or Vision-Language-Action (VLA) transformer-based models~\cite{bousmalis2023robocat, zitkovich2023rt, black2410pi0}.

Further context to the literature in this area is added by works on ROS-based LLM-powered agents such as Bumble~\cite{shah2024bumble}, NavGPT~\cite{zhou2024navgpt}. Other relevant works in this area included non-robotics agents for games such as Minecraft~\cite{wang2023voyager, liu2024odyssey}.

\subsection{ROSBag Analytics / Visualization Frameworks}

The primary method for recording and playing back ROS data is the ROSbag format. While the native tools provide fundamental read/write functionality, the ecosystem has evolved significantly to address the need for more powerful analysis and visualization. 

The MCAP (Modular Container and Playback) format, developed by Foxglove, has emerged as a modern, open-source container file format designed for multimodel timestamped pub/sub or robotics log data. It emphasizes portability, efficient indexing, flexibility in serialization, and is increasingly adopted as an alternative to ROSbag formats. Alongside the format, the Foxglove Studio~\cite{foxglove} is a powerful, extensible visualization and analysis platform that supports both ROS\,1, ROS\,2 and MCAP files. It allows developers to inspect sensor data, plots, and 3D scenes, and exposes APIs and plugins for offline analysis and real-time debugging. Plotjuggler~\cite{plotjuggler} is a lightweight, extensible time-series visualizer widely used in ROS workflows for fast interactive plotting, transform editing, and scripting (Lua).

Heex Technologies offers an event-driven approach to ROSbag data management and analysis, targeting both live and offline ROS workflows. Data is segmented into short MCAP files (e.g., 20s) and evaluated against user-defined / AI-generated triggers (e.g., IMU acceleration exceeding a threshold). Only triggered segments are stored, either from active robots or from existing ROSbags, enabling targeted cloud-based review. This selective-capture model supports efficient ROS data analytics but relies on predefined events, potentially limiting the discovery of unanticipated patterns.

\subsection{MCP Servers and ROS}

An emerging development is the rise of MCP servers as lightweight, modular plugins that equip AI agents with seamless integration into ROS-based systems. A notable extension of MCP is its application in robotics through community-driven projects, openly available on GitHub, such as \textit{ros-mcp} library~\cite{rosmcp}, which bridges AI assistants (e.g., Claude) with installed ROS\,2 applications. This library enables functionalities like ROS resource management (topics, nodes), system interaction (topic publisher, service calls, action goals), simulation control (Gazebo, RViz visualizer), environment debugging and process orchestration, demonstrating a shift from monolithic frameworks (e.g., ROSA, RAI) toward modular, plugin-based architectures. Further expanding MCP's versatility, \textit{ros-mcp-server}~\cite{rosmcpserver}, leverages rosbridge to connect to either ROS\,1 or ROS\,2 systems through WebSockets, and contains specific interfaces to odometry and velocity control topics.  \textit{Auto-mcp}~\cite{auto-mcp} is another custom MCP server enabling real-time discovery of active ROS 2 topics, retrieval of the latest messages, and remote publication of messages to those topics. Additional implementations like the \textit{ros-mcp-server}~\cite{robotmcp} by robotmcp provide comprehensive robot control capabilities, enabling simultaneous control of multiple robots through a unified interface for movement, sound playback, and camera operations.

A recent open-source "ChatGPT for physical data" platform, Bagel~\cite{bagel}, exemplifies how MCP servers can transform analysis of robotic logs. Bagel supports ROS\,1 (.bag), ROS\,2 (.mcap, .db3), PX4 (.ulg), and Ardupilot (.bin) log formats, allowing agents to query logs using natural language and receive metadata summaries, system diagnoses and insights in return. 
Compared to Bagel, we provide more specific tooling, in addition to an in-depth benchmarking.
\section{Implementation}

\subsection{The MCP protocol}

A core capability of modern AI agents is the tool-use module, which allows them to invoke external functions and access structured or unstructured data sources. Most state-of-the-art LLMs are trained to perform such tool calls, enabling them to extend their reasoning and problem-solving abilities beyond their internal knowledge. However, tool calling often requires custom, model-specific integrations, leading to fragmentation and limiting scalability. The MCP addresses these challenges by providing an open interoperability standard that enables seamless, standardized communication between LLMs and external tools or data resources~\cite{anthropic2024mcp}. MCP implements a JSON-RPC client–server architecture, where MCP clients, such as Claude Desktop or Cursor, connect to lightweight MCP servers that securely access both local and remote resources through dedicated one-to-one protocol connections. Additionally, MCP includes an open-source SDK that supports rapid adoption across popular programming languages, including Python.

\subsection{Code architecture / implementation}

The current implementation leverages Python's \texttt{MCP} library to create specialized servers that expose robotics-specific functionality through standardized interfaces. Each MCP server runs as a lightweight process that handles tool registration, parameter validation, and execution orchestration. MCP servers communicate with clients via JSON-RPC messages, enabling a language-agnostic and platform-independent interaction paradigm. The overall architecture of these interactions is summarized in Figure~\ref{fig:diagram}. A modular design is adopted in which each tool is defined in a dedicated function with explicit input and output schemas. These schemas validate user input and provide metadata to MCP clients for generating interactive documentation (see \textit{MCP Tool Schema}). 

FastAPI serves as the underlying web framework, offering automatic request validation, interactive documentation, and robust error handling capabilities that ensure reliable operation during extended analysis sessions.

\subsection{Available tools}
We developed the tools within three main categories, as summarized in Table~\ref{tab:mcp_tools}. Each category is designed to address a specific aspect of robotic data analysis and interaction through the MCP interface.
\subsubsection{Core Data Access And Management}
Robotics experiments generate large, multimodal datasets, and effective analysis begins with robust data management. Accessing, organizing, and filtering recorded trials is essential to ensure that only the most relevant portions of data are examined, particularly when synchronizing control commands with odometry or isolating events of interest. By enabling precise retrieval and reproducible dataset preparation, this category supports efficient systematic evaluation of robot performance.

\subsubsection{Domain-Specific Analysis}
Since robots rely on diverse modalities such as odometry, LiDAR, cameras, and logs, it is necessary to provide dedicated tools for analyzing these common data sources. While basic access to raw messages is possible through core tools, ROSbags can be very large, and extracting meaningful patterns, such as evaluating entire trajectories, can become computationally expensive. A specialized trajectory analysis tool can therefore provide comprehensive statistics, distance measures, and waypoint evaluations far more efficiently. Similarly, coordinate frame transformations are critical for ensuring consistent spatial reasoning across multiple sensors, and a dedicated TF tree analysis tool ensures that these relationships remain coherent and interpretable. By separating these domain-focused tasks into specialized tools, the analysis process becomes both computationally efficient and directly aligned with the unique requirements of robotic systems.

\begin{tcolorbox}[mybox, title={MCP Tool Schema: \texttt{get\_messages\_in\_range}}, label={lst:get-messages-schema}]
\textbf{Description:} Get all messages from a topic within a time range.  

\textbf{Input Schema:}
\begin{lstlisting}[style=json, basicstyle=\scriptsize\ttfamily]
{
  "type": "object",
  "properties": {
    "topic": {
      "type": "string",
      "description": "ROS topic name"
    },
    "start_time": {
      "type": "number",
      "description": "Start unix timestamp in seconds"
    },
    "end_time": {
      "type": "number",
      "description": "End unix timestamp in seconds"
    },
    "max_messages": {
      "type": "integer",
      "description": "Maximum messages to return (default: 100)"
    },
    "bag_path": {
      "type": "string",
      "description": "Optional: specific bag file or directory to search"
    }
  },
  "required": [
    "topic",
    "start_time",
    "end_time"
  ]
}
\end{lstlisting}
\end{tcolorbox}

\subsubsection{Visualization and Plotting}
Visualization bridges the gap between raw data and interpretable insights, making system behavior immediately apparent. Time-series plots expose dynamic trends such as responsiveness or drift, while spatial trajectory views highlight path following quality and navigation efficiency. Modality-specific plots, such as sensor visualizations, further help validate perception outputs.

\begin{figure}
    \centering
    \includegraphics[width=\linewidth]{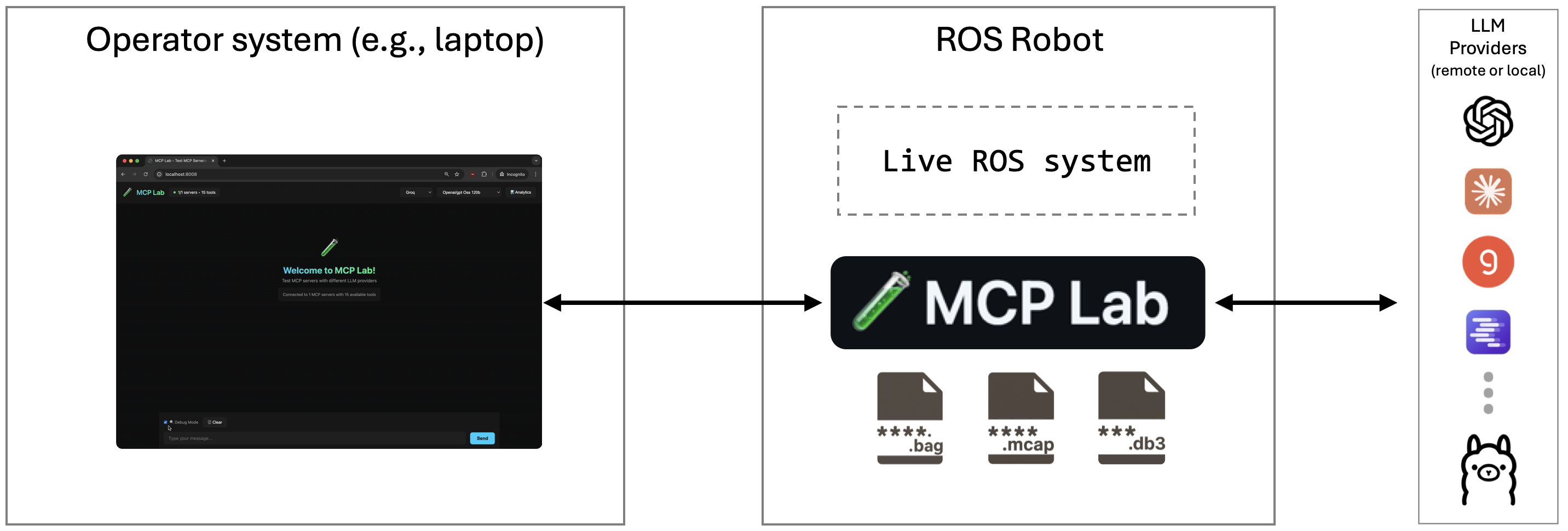}
    \caption{System architecture of MCP Lab with remote connection support, illustrating the integration of MCP servers and multiple LLM providers.}
    \label{fig:mcp-lab-robot-setup}
\end{figure}

\subsection{MCP Lab}
    
MCP Lab~\cite{mcplab} is a modular testing environment for MCP servers that supports multiple LLM providers. Its primary purpose is to provide a controlled and reproducible setup for benchmarking and analysis. To achieve this, MCP lab standardizes access to ROSbag repositories, manages tool invocation, and records structured metrics throughout the evaluation process. These features enable systematic comparisons between models by allowing us to measure tool usage behavior, task completion rates, and overall performance under consistent conditions. In addition, MCP Lab serves as the accompanying framework for this work, ensuring that the experimental results presented in this paper are both replicable and grounded in a unified testing methodology.
Figure~\ref{fig:mcp-lab-robot-setup} shows the MCP Lab interface connected to different LLM providers, enabling tool testing and performance tracking of MCP servers without local ROS installation.

\begin{figure*}[t] 
    \centering
    \begin{subfigure}{.48\textwidth}
        \includegraphics[width=\textwidth,  height=0.4\textheight]{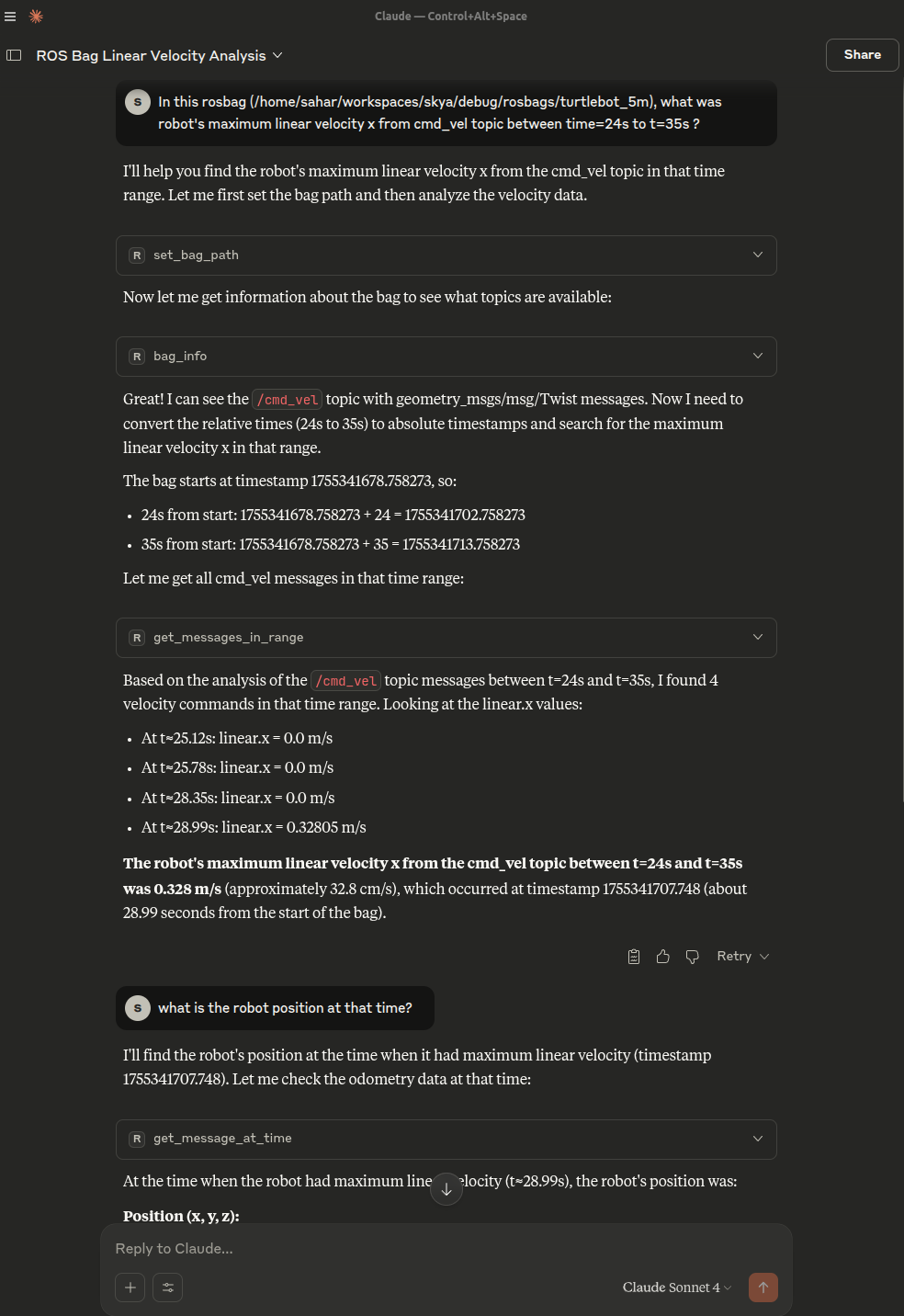}
        \caption{Sample conversation with Claude Desktop.}
        \label{fig:claude}
    \end{subfigure}%
    \hfill
    \begin{subfigure}{.48\textwidth}
        \includegraphics[width=\textwidth, height=0.4\textheight]{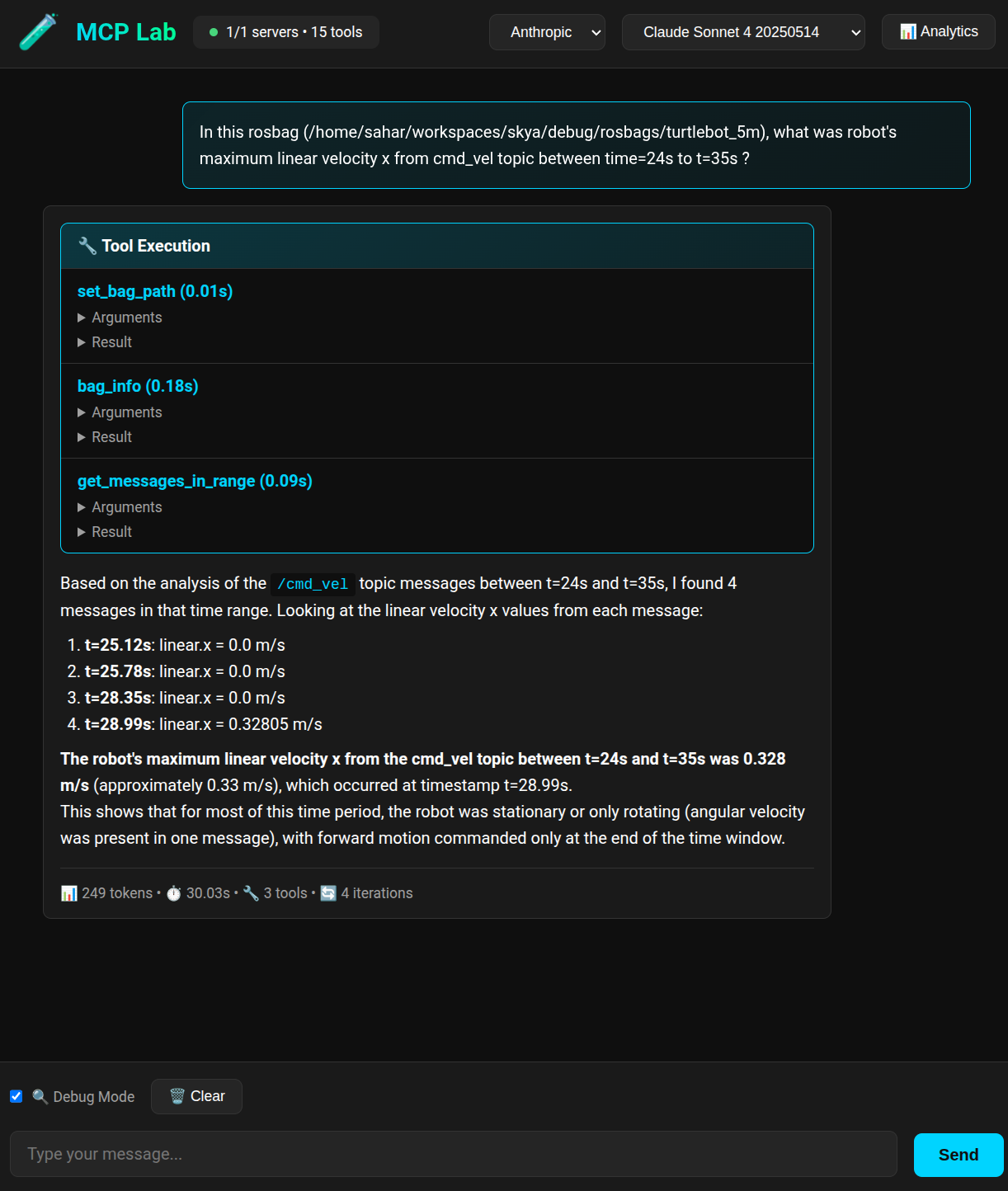}
        \caption{Sample conversation in our custom MCP-lab web interface.}
        \label{fig:interface}
    \end{subfigure}%
    \caption{Conversation workflow showing the MCP tool interaction within the Claude Desktop App and MCP-Lab interfaces.}
    \label{fig:conversation}
\end{figure*}

\begin{sidewaystable*}
\centering

\caption{Description of tools in the initial release of the Rosbags MCP server.}
\label{tab:mcp_tools}
\begin{tabular}{@{}l l l l@{}}
\toprule
\textbf{Category} & \textbf{Task Type} & \textbf{Tool} & \textbf{Description} \\
\midrule
\multirow{7}{*}{Core Data Access \& Management} 
  & \multirow{3}{*}{Bag Operations} 
    & \texttt{B1: set\_bag\_path} & Set the path to a rosbag file or directory \\
  &  & \texttt{B2: list\_bags} & List all available rosbag files in the directory \\
  &  & \texttt{B3: bag\_info} & Retrieve bag metadata: topics, counts, duration \\
\cmidrule(lr){2-4}
  & \multirow{3}{*}{Message Retrieval} 
    & \texttt{M1: get\_message\_at\_time} & Get message from a topic at a specific timestamp \\
  &  & \texttt{M2: get\_messages\_in\_range} & Get all messages from a topic within a time range \\
  &  & \texttt{M3: search\_messages} & Search messages using conditions (regex, equals, etc.) \\
\cmidrule(lr){2-4}
  & Data Filtering 
    & \texttt{FB: filter\_bag} & Create a filtered copy of a bag file by topic, time, or rate \\
\midrule
\multirow{5}{*}{Domain-Specific Analysis} 
  & Trajectory Analysis 
    & \texttt{AT: analyze\_trajectory} & Compute trajectory metrics (distance, speed, waypoints) \\
  & LiDAR Analysis 
    & \texttt{AS: analyze\_lidar\_scan} & Analyze LiDAR scans for obstacles, gaps, statistics \\
  & Log Analysis 
    & \texttt{AL: analyze\_logs} & Parse/analyze ROS logs; filter by level or node \\
  & Transform Analysis 
    & \texttt{GT: get\_tf\_tree} & Get TF tree of coordinate frame relationships \\
  & Image Analysis 
    & \texttt{GI: get\_image\_at\_time} & Extract camera image at specific time (base64 JPEG) \\
\midrule
\multirow{3}{*}{Visualization \& Plotting} 
  & Time Series Visualization 
    & \texttt{P1: plot\_timeseries} & Plot time series data with multiple styles \\
  & Spatial Visualization 
    & \texttt{P2: plot\_2d} & Create 2D trajectory plots (XY) \\
  &  & \texttt{P3: plot\_lidar\_scan} & Visualize LiDAR scans as polar plots \\
\bottomrule
\end{tabular}

\vspace{2em}

\caption{Task completion and tool usage across different models.}
\label{tab:tool_usage}

\renewcommand{\arraystretch}{1.5}

\footnotesize{%
\begin{tabular}{@{}l*{9}{cc}@{}}
\toprule
\textbf{\normalsize Task} & \textbf{Required} &
\textbf{\scriptsize Sonnet} & 
\textbf{\scriptsize Sonnet} & 
\textbf{\scriptsize GPT-4o} & 
\textbf{\scriptsize GPT-5} & 
\textbf{\scriptsize Qwen3} & 
\textbf{\scriptsize Llama 4} & 
\textbf{\scriptsize Kimi} & 
\textbf{\scriptsize GPT OSS} \\
 & \textbf{tools} &
\textbf{4} & 
\textbf{3.7} & 
\textbf{Mini} & 
\textbf{Mini} & 
\textbf{32B} & 
\textbf{Scout} & 
\textbf{K2} & 
\textbf{20B} \\
\midrule

{\scriptsize What bags do you have in \code{/path/to/folder}}
& B2 & \ding{51} &\ding{51} + B1 &\ding{51} &\ding{51} &\ding{51} &\ding{51} &\ding{51} &\ding{51} \\

{\scriptsize Plot the velocity (linear and angular) from \code{t=Xs} to \code{t=Ys}}
& P1 & \ding{51} & \ding{55} & \ding{55} & \ding{51} &\ding{51} & \ding{55} & \ding{51} & \ding{51} \\

{\scriptsize Generate a trajectory summary for the entire bag: total distance, mean/max speeds} 
& AT&\ding{51} &\ding{51} &\ding{51} &\ding{51} &\ding{51} &\ding{55} &\ding{51} & \ding{51}\\

{\scriptsize Plot the trajectory of the robot } 
& P2 &\ding{51} &\ding{51} + AT &\ding{51} + AT&\ding{51} &\ding{51} &\ding{55} &\ding{51} + AT &\ding{51} + AT, AS \\

{\scriptsize Has the robot ever passed close by this position: (\code{x=X, y=Y}), within \code{Z} meters? } 
& M3 &\ding{51} &\ding{55} &\ding{51}+ M2 & \ding{51} + M2&\ding{51} &\ding{55} &\ding{51} & \ding{51}\\

{\scriptsize Does a lidar indicate any obstacle? When is the first time the robot sees an obstacle within \code{X}} & AS, M3
& \ding{51}& \ding{55}& \ding{55}& \ding{55}& \ding{55}&\ding{55} &\ding{51} &\ding{55} \\

{\scriptsize Plot the lidar scan when \code{t=Xs} and plot a histogram of the \code{cmd\_vel linear x}} & P1, P3 & \ding{51}& \ding{55}& \ding{55}& \ding{51} & \ding{51} + M1 & \ding{55}& \ding{51}& \ding{51} \\

{\scriptsize List all the frames and their relationships }  & GT 
&\ding{51} &\ding{51} &\ding{51} &\ding{51} &\ding{55} &\ding{55} &\ding{51} &\ding{55} \\

{\scriptsize Filter the rosbag and create a new rosbag with only \code{cmd\_vel} and \code{odom}} 
&FB &\ding{51} &\ding{51} &\ding{51} &\ding{51} &\ding{51} &\ding{55} &\ding{51} & \ding{51}\\

{\scriptsize Robot's maximum commanded linear vel. between \code{X-Ys}, time and position when it occurs?}
& M2, M1 & \ding{51} & \ding{51}&\ding{55} &\ding{51} & \ding{55}&\ding{55} &\ding{51} + B3,M2,M3 & \ding{55}\\

\bottomrule
\end{tabular}%
}

\end{sidewaystable*}

\section{Experimental Results}

To assess the effectiveness of our MCP-based robotic analysis framework, we conducted a comprehensive evaluation across multiple language models using standardized set of robotics data analysis tasks, ranging from basic data retrieval operations (e.g. extract sensor readings) to complex analytical workflows that require multiple tools. 

For illustrative purposes of the capabilities of the proposed MCP server, the following is an example conversation below.

\begin{tcolorbox}[mybox, title={Examples of MCP Tool Requests},label={lst:mcp-examples}, fontupper=\small]

\textbf{User:} Has the robot ever passed close by this position: (x=2, y=-2), within 0.5 meters?  

{MCP \texttt{search\_message} Request:}
\begin{lstlisting}[style=json, basicstyle=\scriptsize\ttfamily]
{
  "bag_path": "/path/rosbags/turtlebot_5m",
  "topic": "/odom",
  "condition_type": "near_position",
  "value": "2,-2,0.5",
  "limit": 10
}
\end{lstlisting}
\textbf{Agent:} The robot indeed passed within 0.5 m of the target position (2, –2). Its closest approach was 0.217 m (21.7 cm) at position (1.793, –1.933), recorded at timestamp 1755341898.5013318.

\hrulefill 

\vspace{0.5em}

\textbf{User:} During the first 30 seconds, when the robot has commanded an angular velocity greater than 0.4?  

{MCP \texttt{get\_message\_in\_range} Tool Request:}

\begin{lstlisting}[style=json, basicstyle=\scriptsize\ttfamily]
{
  "topic": "/cmd_vel",
  "end_time": 1755341708.758273,
  "start_time": 1755341678.758273,
  "max_messages": 50
}
\end{lstlisting}
\textbf{Agent:} During the first 30 seconds, the robot had angular velocity greater than 0.4 rad/s at 3 times (6.19s, 16.61s, 25.78s), all with angular velocity of 0.656 rad/s.  

\hrulefill 

\vspace{0.5em}

\textbf{User:} What range of x and y has the robot traveled?  

{MCP \texttt{analyze\_trajectory} Tool Request:}
\begin{lstlisting}[style=json, basicstyle=\scriptsize\ttfamily]
{
  "bag_path": "/path/rosbags/turtlebot_5m",
  "pose_topic": "/odom",
  "start_time": 1755341678.758273,
  "end_time": 1755341985.7409644,
  "include_waypoints": false
}
\end{lstlisting}
\textbf{Agent:} The robot traveled within:\\
X range: -2.074  to 2.168  meters (total span: 4.243 meters)\\
Y range: -2.017 to 2.000  meters (total span: 4.017 meters)

\hrulefill 

\vspace{0.5em}

\textbf{User:} Plot the commanded and actual velocities, linear and angular, for the first 30 seconds  

{MCP \texttt{plot\_timeseries} Tool Request:}

\begin{lstlisting}[style=json, basicstyle=\scriptsize\ttfamily]
{
  "fields": [
    "cmd_vel.linear.x",
    "odom.twist.twist.linear.x",
    "cmd_vel.angular.z",
    "odom.twist.twist.angular.z"
  ],
  "start_time": 1755341678.758273,
  "end_time": 1755341708.758273,
  "title": "Commanded vs Actual Velocities (First 30 seconds)",
  "x_label": "Time (s)",
  "y_label": "Velocity"
}
\end{lstlisting}
\textbf{Agent:} I've created a time series plot showing both the commanded and actual velocities for the first 30 seconds of the ROSbag.
\begin{center}
  \includegraphics[width=0.9\columnwidth]{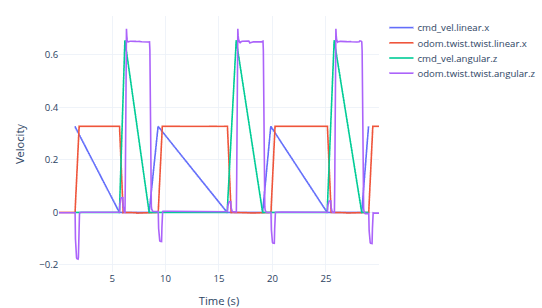}
\end{center}

\end{tcolorbox}

\begin{figure*}[t]
    \centering
    \setlength{\figurewidth}{.95\textwidth}
    \setlength{\figureheight}{.4\textwidth}
    \input{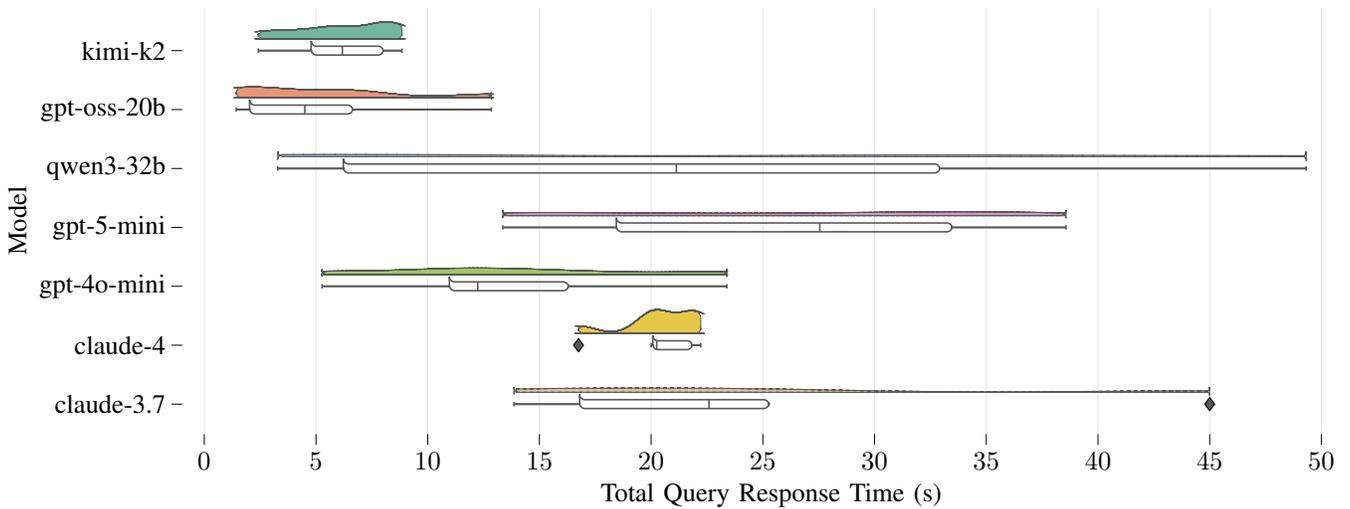}
    \caption{Response time distribution per model across successful tool-calling tasks. Lower and more consistent response time indicates superior computational efficiency and more reliable decision-making processes.}
    \label{fig:raincloud-latency}
\end{figure*}

\begin{figure*}[t]
    \centering
    \setlength{\figurewidth}{.95\textwidth}
    \setlength{\figureheight}{.4\textwidth}
    \input{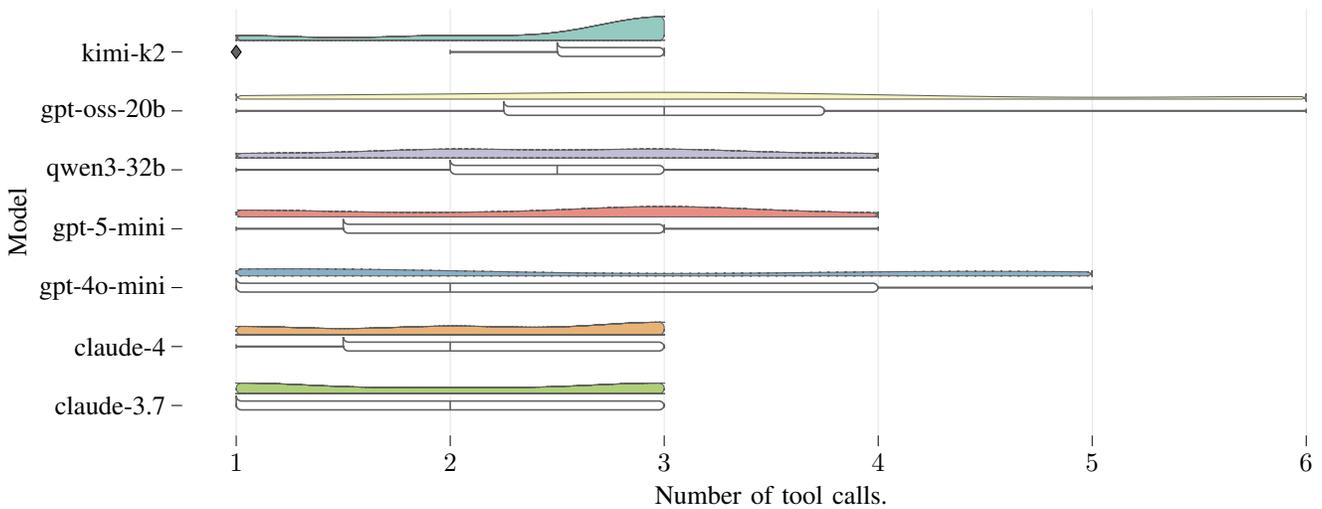}
    \caption{Distribution of number of tools used per successful response across different models. The visualization reveals how many tools each model requires to complete identical tasks, with lower values indicating more efficient problem-solving approaches.}
    \label{fig:raincloud-tools}
\end{figure*}

\subsection{Experiment Setup}
The evaluation was performed using eight different language models, including GPT-4o, Claude 3.5 Sonnet, and several open-source models. All experiments were conducted within MCP Lab~\cite{mcplab}, see Figure~\ref{fig:conversation} for an example conversation.

Each model was given identical task prompts, with access to the same ROSbag data and analysis utilities. A task was considered successful if the model called the correct tools and produced valid results without manual intervention. Partial success was recorded if the reasoning process was correct but required human guidance or additional tool calls beyond those baseline tools.

\subsection{Model Performance Analysis}

Performance varied considerably across the tested models. Table~\ref{tab:tool_usage} provides a detailed comparison of task completion and tool usage. The required tools column indicates the baseline tools that should be invoked for each query.
Among the tested models, Claude Sonnet 4 and Kimi K2 demonstrated the most consistent performance, successfully completed all the tasks, followed closely by GPT-5 mini with 9 successful completions (90\% success rate). These frontier models show particular strength in handling complex tasks that required combining multiple tool calls. Both Qwen3 and GPT OSS completed 7 tasks (70\% success rate), struggling on the search specific data and frame transformations. In contrast, Sonnet 3.7, GPT-40 mini and Llama 4, as smaller models, showed notable limitations. These models frequently failed to complete multi-step analytical tasks, often producing incomplete results or requiring extensive guidance on tool calling.

In addition, we evaluated the models based on their end-to-end response time and the number of tool calls required to complete the tasks presented in Table~\ref{tab:tool_usage}. Figure~\ref{fig:raincloud-latency} reveals the response time consistency across models. Kimi-k2 and Claude-4 demonstrate the most stable and predictable response times, with tight distributions around low median values and minimal variance, indicating reliable and efficient tool-calling behavior. Conversely, smaller models exhibit substantially higher response time variability, characterized by broader distributions. This increased response time and inconsistency likely reflects suboptimal decision-making processes, where these models may take unnecessary computational approaches rather than efficiently utilizing available tools.
Figure~\ref{fig:raincloud-tools} shows that Kimi-k2, Claude-4, and Claude-3.7 exhibit superior performance with the most efficient tool usage patterns, consistently requiring fewer tools to complete the same tasks. In contrast, GPT models and other open-source models demonstrate less efficient behavior, frequently invoking more tools than necessary for task completion.

\section{Discussion \& Conclusion}

This paper introduced an MCP server for ROS and ROS 2 bag analysis, enabling natural language interaction with robotic datasets through LLMs and VLMs. Our implementation provides domain-specific tooling for trajectory analysis, laser scan processing, coordinate frame transformations, and time series visualization, bridging the gap between complex robotic data and conversational AI interfaces. The experimental evaluation across eight state-of-the-art models revealed substantial disparities in tool calling capabilities, with Kimi K2 and Claude Sonnet 4 achieving 100\% task completion rates, while smaller models struggled with multi-step analytical workflows. These results demonstrate that while MCP offers a standardized protocol for tool integration, model selection remains critical for effective deployment in agentic embodied AI applications.

Our final results emerge from extensive iterative refinement of tool descriptions and interfaces. Through this process, we identified several factors that significantly influence tool calling success rates: schema clarity directly impacts invocation accuracy, with ambiguous descriptions leading to malformed requests; argument complexity poses challenges for smaller models, which frequently fail to construct valid parameter sets; context preservation across sequential tool calls proves problematic, as models often forget previously established parameters like folder paths; and perhaps most critically, the total number of available tools inversely correlates with success rates, particularly when tools share similar functionalities or overlapping parameter spaces. These observations suggest that optimal MCP server design requires careful balance between functionality breadth and interface simplicity.

Future work will focus on systematic ablation studies to dissect successful tool design patterns and establish best practices for MCP-based robotic systems. Specifically, we plan to investigate schema optimization strategies, develop context-aware tool chaining mechanisms, and explore adaptive tool exposure based on model capabilities. Additionally, extending the framework to support real-time robot control and multi-robot coordination presents compelling opportunities for advancing agentic embodied AI applications. The open-source release of our implementation aims to accelerate community-driven development of natural language interfaces for robotics, ultimately making robotic data analysis more accessible to non-expert users while maintaining the precision required for professional deployment.

\bibliographystyle{IEEEtran}
\bibliography{bibliography}

\end{document}